%% file: acl_latex.tex
\pdfoutput=1

\documentclass[11pt]{article}
\usepackage{authblk}

\usepackage[preprint]{acl}

\usepackage{times}
\usepackage{latexsym}

\usepackage[T1]{fontenc}

\usepackage[utf8]{inputenc}

\usepackage{microtype}

\usepackage{inconsolata}

\usepackage{graphicx}
\usepackage{multirow}
\usepackage{paralist}
\usepackage{colortbl}
\usepackage{bm} 
\usepackage{subfigure}
\usepackage{amsmath}
\usepackage{amsfonts}
\usepackage{booktabs} 
\usepackage[ruled,vlined]{algorithm2e}
\SetAlFnt{\small}  

\title{SECURA: Sigmoid-Enhanced CUR Decomposition with Uninterrupted Retention and Low-Rank Adaptation in Large Language Models
}

\author[1,2]{Yuxuan Zhang
\thanks{Corresponding author. Email: \texttt{MeCuping@outlook.com}. 
\newline \indent GitHub repository: 
\newline \indent \url{https://github.com/MeCuping/SECURA}}}

\affil[1]{Aberdeen Institute of Data Science and Artificial Intelligence, South China Normal University}
\affil[2]{Department of Computing Science, University of Aberdeen}

\begin{document}
\maketitle
\begin{abstract}
With the rapid development of large language models (LLMs), fully fine-tuning(FT) these models is becoming increasingly infeasible due to high computational demands. Moreover, FT also increases the risk of catastrophic forgetting. As an alternative, Low-Rank Adaptation (LoRA) has been proposed. By fine-tuning only a small subset of parameters, LoRA achieves performance similar to FT while significantly reducing resource requirements. However, since LoRA inherits FT’s design, the issue of catastrophic forgetting still remains.
To address these limitations, we propose \textbf{SECURA}: \textbf{S}igmoid-\textbf{E}nhanced \textbf{CU}R Decomposition Lo\textbf{RA}, a novel PEFT variant designed to mitigate catastrophic forgetting while improving fine-tuning performance. Our method introduces a novel normalization technique, \textbf{Sigmoid based Magnitude Norm(S-MagNorm)},  which enhances parameter retention and improve fine-tuning performance. SECURA has been evaluated on a diverse range of tasks, including mathematical problem-solving (GSM8K), complex question-answering (CNNDM), translation (NewsDE), and complex multiple-choice reasoning (LogiQA). Experimental results demonstrate that it achieves an average fine-tuning improvement of \textbf{3.59\%} across 4 MCQ sphere tasks and a \textbf{2.51\%} improvement in 5 QA sphere tasks across Gemma2 2b, Qwen2 1.5b, Qwen2 7b,Llama3 8b,Llama3.1 8b compared to DoRA. Additionally, SECURA demonstrates superior knowledge retention capabilities, achieves state-of-the-art performance in 16 continual learning tests maintaining more than \textbf{70\%}  accuracy on LLMs' basic knowledge compared with Experience Replay(ER), sequential learning(SEQ),EWC, I-LoRA and CUR-LoRA.
\end{abstract}

\input{sections/introduction}
\input{sections/background}
\input{sections/methdology}
\input{sections/experiment}

\input{sections/discussion}

\section{Conclusion}
We propose SECURA, a novel PEFT fine-tune method that integrates our newly proposed S-MagNorm normalization and CABR decomposition to address catastrophic forgetting in LLM fine-tuning. SECURA dynamically balances parameter updates using a Sigmoid-based pruning mechanism, preserving critical knowledge while adapting to new tasks. Experiments on 18 datasets and 5 LLMs show SECURA outperforms standard LoRA, with a 3.63\% improvement on MCQ tasks and 2.56\% on QA tasks, while retaining over 70\% of base knowledge in continual learning scenarios. Paving the way for sustainable and ethical deployment of large AI models.

\section{Limitations}

\begin{itemize}
    \item \textbf{Lack of experiments on 70B+ LLMs:} Due to device limitations, further research on larger models (70B+ parameters) could not be conducted. Future work will explore scaling SECURA to more massive LLMs.

    \item \textbf{Computational Overhead:} S-MagNorm introduces additional matrix operations during training, increasing per-step time by approximately 1.18\% compared to LoRA. Future work could focus on optimizing the normalization layer for real-time applications.

\end{itemize}

\bibliography{custom}

\appendix

\input{sections/appendix}

\end{document}

%% file: sections/introduction.tex
\section{Introduction}
With the rapid development of transformer-based generative models, parameter scales have grown exponentially to enhance multi-task capabilities. Notable examples in natural language models(NLP) like LLaMA 3.1~\citep {dubey2024llama} and Qwen 2.5~\citep {qwen2.5} achieve remarkable performance on complex tasks including text generation, logical reasoning, and multilingual processing.  However, their massive parameterization (e.g., 70B+ parameters) creates deployment barriers: Full fine-tuning requires over 1.5TB of GPU memory and weeks of computation. Even with distilled versions, they remain prohibitively expensive for most individual developers and small research labs.

Parameter-Efficient Fine-Tuning (PEFT)~\citep {xu2023parameter} methods, such as Low-Rank Adaptation (LoRA)~\citep {whitehouse-etal-2024-low}, address this challenge. LoRA inject trainable tiny low-rank matrices (typically <0.1\% of original parameters) and frozen basic weights to reduce the computational cost of fine-tuning. Performed well in multi-task learning, multilingual summarisation, and transfer learning~\citep{whitehouse-etal-2024-low,zhao2024adamergex}. 

However, LoRA remains an approximation of FT, meaning it can hardly match or exceed the performance of full fine-tuning. Furthermore, it also inherits FT's critical issue: catastrophic forgetting~\citep {MCCLOSKEY1989109} . While freezing the base model weights alleviates this problem to some extent, the addition of LoRA matrices reintroduces the issue. Additionally, fine-tuning fewer parameters, while reducing computational cost, increases the risk of catastrophic forgetting before the model finds an optimal solution. Although weight normalization techniques (Salimans and Kingma, 2016) can be used to mitigate these issues, they do not provide a comprehensive solution. Existing LoRA variants primarily focus on improving specific aspects: enhancing performance (e.g. DoRA~\citep{liu2024doraweightdecomposedlowrankadaptation}, AdaLoRA~\citep{zhang2023adaloraadaptivebudgetallocation}), accelerating convergence (e.g., PiSSA~\citep{meng2024pissaprincipalsingularvalues}), or only reducing catastrophic forgetting (e.g. I-LoRA~\citep{ren2024analyzingreducingcatastrophicforgetting}, CUR-LoRA~\citep{https://doi.org/10.5281/zenodo.12730055}). However, few approaches simultaneously address uninterrupted retention of knowledge and keep the performance.

Motivated by challenges in catastrophic forgetting, we drew inspiration from QKV pruning theory~\citep{lv2024kvprunerstructuralpruningfaster}, and Elastic Weight Consolidation(EWC)~\citep{aich2021elasticweightconsolidationewc} to design a new LoRA variant: \textbf{SECURA}. Our approach leverages two core innovations:

\begin{compactitem}
    \item \textbf{CABR Decomposition}: To enhance the low performance of CUR-LoRA, we introduced an inverse low-rank adaptation matrix into the trainable $U$-matrix. By incorporating additional dimensions, CABR improves CUR-LoRA's performance, achieving results comparable to or exceeding those of standard LoRA.
    \item \textbf{S-MagNorm Normalization}: To balance model stability and performance, we designed S-MagNorm—a normalization method using Sigmoid's gradual transition property. It dynamically adjusts parameters during training to prevent catastrophic forgetting by prioritizing the collective impact of central parameter changes over extreme values, eliminating the need for historical data or prior outputs. This ensures balanced weight updates while maintaining adaptability.
    \newline
\end{compactitem}

By integrating S-MagNorm and CABR decomposition, SECURA effectively addresses memory retention challenges in continual learning while surpassing existing LoRA methods and their variants in accuracy under comparable computational budgets. Experiments on 5 LLMs (Gemma2 2B, Qwen2 1.5B/7B, Llama3 8B, Llama3.1 8B) demonstrate SECURA's superiority: it achieves \textbf{3.63\%} average improvement on 4 multiple-choice (MCQ) tasks and \textbf{2.56\%} on 5 question-answering (QA) tasks over baseline LoRA, even rivaling advanced variants like DoRA. For knowledge retention, SECURA maintains over \textbf{70\%} accuracy on fine-tuned test sets, outperforming state-of-the-art methods such as Experience Replay (ER)~\citep{fedus2020revisitingfundamentalsexperiencereplay}, sequential learning (SEQ)~\citep{sutskever2014sequencesequencelearningneural}, EWC~\citep{aich2021elasticweightconsolidationewc},I-LoRA~\citep{ren2024analyzingreducingcatastrophicforgetting} and CUR-LoRA~\citep{https://doi.org/10.5281/zenodo.12730055}. These results highlight SECURA's dual strengths in task adaptation and preserving pre-trained knowledge.

%% file: sections/background.tex
\section{Background}

\begin{figure*}[tb]
    \centering
    \includegraphics[width=\textwidth]{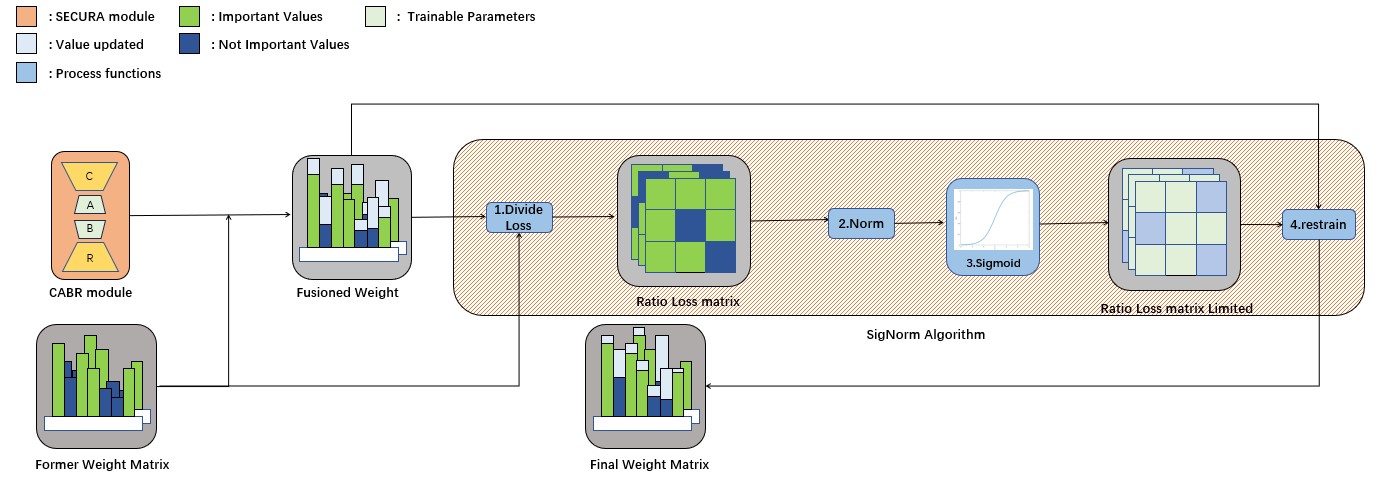}
    \caption{S-MagNorm Update method: The figure illustrates the process flow of the S-MagNorm Normalization algorithm. Shows the steps of starting with the fusion of the former weight matrix with the CABR module and moving through the ratio loss matrix calculation, the normalization and sigmoid steps, followed by the final limiting of the ratio loss matrix.}
    \label{fig:S-MagNorm_Update}
    \vspace{-1em}
\end{figure*}

\paragraph{Low-Rank Adaption.}LoRA~\citep {whitehouse-etal-2024-low} is a method that uses low-rank matrices, hypothesizing that in the parameters matrix, only few parameters controls main knowledge. By introducing low-rank parameter matrices A and B, “it can generate a matrix that provides the same output, where A initialized by random sampling, B initialized by zeros.

Formally, consider a weight matrix \( \boldsymbol{W} \in \mathbb{R}^{h \times d} \) within the original LLMs. LoRA introduces two low-rank matrices, \( \boldsymbol{A} \in \mathbb{R}^{h \times r} \) and \( \boldsymbol{B} \in \mathbb{R}^{r \times d} \), where \( r \ll \min(h, d) \). Instead of directly updating the weight matrix, LoRA modifies the model's forward pass according to the following equation:
\[
    \boldsymbol{W}' = \boldsymbol{o} + \Delta \boldsymbol{o} = \boldsymbol{W} + \boldsymbol{AB}
\]
By freezing the basic weight and adding the AB matrix the model can be fine-tuned similarly like FT. 

\paragraph{CUR-Decomposition and Its Application in CUR-LoRA.}
CUR-Decomposition~\citep{hamm2019curdecompositionsapproximationsperturbations} is a matrix factorization method that selects specific rows and columns from a matrix. It is used to decompose a matrix in various ways depending on the design. When combined with gradient descent, it can be applied easily using the following formula:
\[
    \boldsymbol{W} = \boldsymbol{C} \boldsymbol{U} \boldsymbol{R}(\boldsymbol{x}),
\]
where \( \boldsymbol{W} \in \mathbb{R}^{h \times d} \) is the original parameter weight, \( \boldsymbol{C} \in \mathbb{R}^{h \times r} \) and \( \boldsymbol{R} \in \mathbb{R}^{r \times d} \) are the selected columns and rows from \( \boldsymbol{W} \), and \( \boldsymbol{U} \in \mathbb{R}^{r \times r} \) is a learnable matrix. Here, \( r \ll \min(h, d) \).

CUR-LoRA~\citep{https://doi.org/10.5281/zenodo.12730055} builds on CUR-Decomposition, by selecting the least important rows and columns as \( \boldsymbol{C} \) and \( \boldsymbol{R} \). These selected rows and columns are much fewer than the original matrix and have the lowest norms, which leads to slower updates. This decomposition allows CUR-LoRA to focus on updating the less important parts of the matrix, because less important parameters are more sensitive to slower updates, while important ones are less affected. In this way, the model can preserve the previously learned knowledge while efficiently learning new information, improving both training efficiency and performance.

\paragraph{Dynamic Network Pruning.}  
Dynamic Network Pruning~\citep{lin2020dynamicmodelpruningfeedback}  is an input-adaptive method that improves the efficiency of neural networks by dynamically adjusting their structure during inference. Unlike static pruning~\citep{siciliano2024staticpruningdenseretrieval}, which removes redundant weights or neurons, dynamic pruning adapts the network based on the characteristics of each input. This approach allows the network to selectively activate relevant parameters, optimizing computational cost.

Formally, consider a weight matrix \( \boldsymbol{W} \in \mathbb{R}^{h \times d} \) and a binary mask matrix \( \boldsymbol{M}(\boldsymbol{x}) \in \{0, 1\}^{h \times d} \), where \( \boldsymbol{x} \) represents the input data. Dynamic pruning modifies the forward pass as follows:
\[
    \boldsymbol{y} = \sigma \left( (\boldsymbol{W} \odot \boldsymbol{M}(\boldsymbol{x})) \boldsymbol{x} \right)
\]
where \( \odot \) denotes element-wise multiplication, and \( \sigma \) represents the activation function.

The mask \( \boldsymbol{M}(\boldsymbol{x}) \) is dynamically generated based on input-dependent criteria such as weight importance, gradient magnitudes, or gating mechanisms. This ensures that only a subset of parameters is used for each input.

Dynamic pruning offers significant computational efficiency by reducing unnecessary operations, allowing the network to adjust its complexity based on input difficulty. However, generating \( \boldsymbol{M}(\boldsymbol{x}) \) and ensuring efficient real-time pruning remain active research challenges.

%% file: sections/methdology.tex
\section{Methodology}
Our proposed SECURA method is based on three key components: CABR-LoRA inverse decomposition fine-tuning, the S-MagNorm Normalization algorithm, and two types of SECURA Merge method. Drawing inspiration from dynamic network pruning and CUR-decomposition, we hypothesize that parameters with higher norms in the original model are more important, as they store more critical knowledge. By constraining these important parameters and focusing on fine-tuning the less important ones, SECURA can achieve remarkable performance while significantly mitigating catastrophic forgetting. Appendix~\ref{app:Pseudo} shows all intuitive equation derivation. 

\begin{figure*}[tb]
    \centering
    \includegraphics[width=\textwidth]{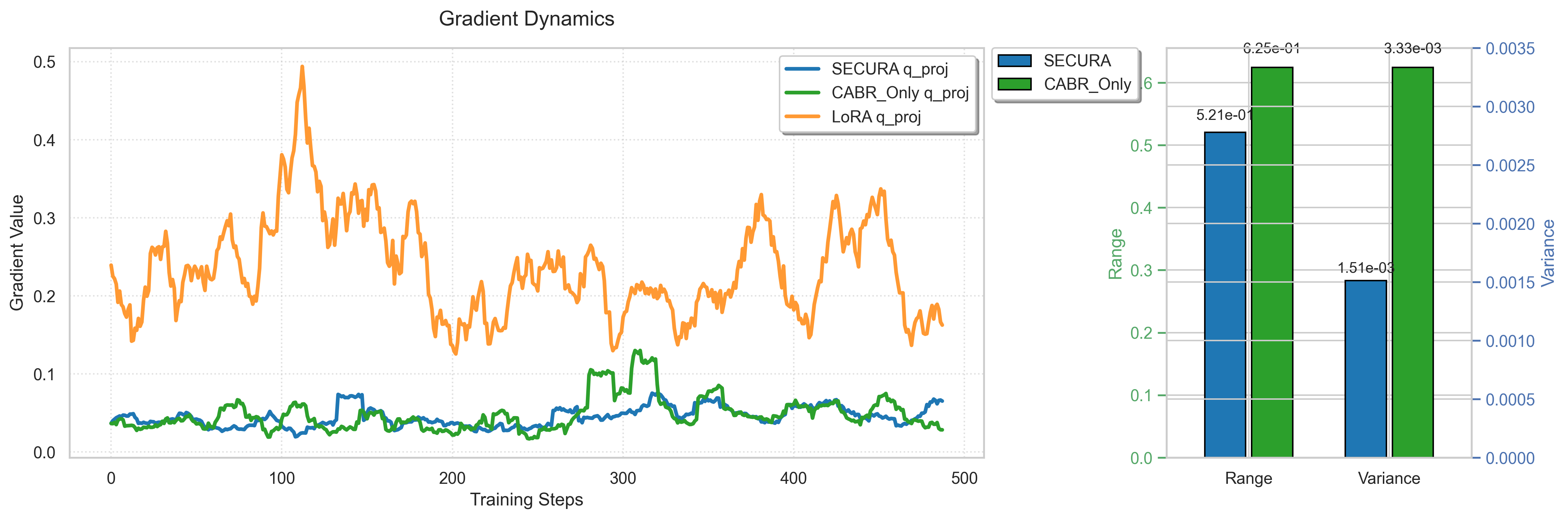}
    \caption{Gradient Analysis: The gradient variations during training shows that LoRA exhibits higher fluctuations, indicating greater parameter reshaping, which may increase the risk of catastrophic forgetting. In contrast, SECURA (CABR-LoRA + S-MagNorm) and CABR-LoRA Only shows lower gradient changes,  avoided excessive drift, suggesting more stable parameter updates, improving catastrophic forgetting mitigation and ability of finding a better optimum for the model's parameters. Additionally, the range and variance comparsion shows the essential role of S-MagNorm. Hyperparameter settings are detailed in Appendix \ref{app:hyper_param}.}
    \label{fig:Grad_analysis}
    \vspace{-1em}
\end{figure*}

\subsection{CABR-LoRA Inverse Decomposition Initialization}

CABR-LoRA is based on CUR-LoRA, utilizing CUR decomposition to break down the weight matrices. As shown in Figure \ref{fig:S-MagNorm_Update}, CABR-LoRA is initialized using a inverse low-rank decomposition of the 
\(U\) matrix. This allows the "less important" columns and rows to remain smaller while maintaining a higher-dimensional space for fine-tuning SECURA. Where \( \mathcal{W}_{\text{CABRWeight}} \) is the initialized CABR-LoRA weight, \( C \) and \( R \) represent the less important columns and rows, while \( A_{\text{initial}} \) and \( B_{\text{zeros}} \) are the trainable matrices with weights \( \mathcal{W}_{\text{A}} \) and \( \mathcal{W}_{\text{B}} \). The initialization formula for SECURA is given as follows:
\begin{equation}
    \mathcal{W}_{\text{CABRWeight}} = C\cdot A_{\text{initial}}\cdot B_{\text{zeros}}\cdot R.
\end{equation}

CUR-LoRA reduces the number of trainable parameters to locate least important \( \boldsymbol{C} \) and \( \boldsymbol{R} \), but this often leads to the loss of important parameters, causing performance degradation. To address this, we introduce CABR decomposition, which decomposes the \( U \) matrix, into two matrices: \( \mathcal{W}_{\text{A}} \) and \( \mathcal{W}_{\text{B}} \), which shapes are \( r \times m \) and \( m \times r \). Respectively here \( m > r \). \( \mathcal{W}_{\text{A}} \) is initialized via SVD decomposition, while \( \mathcal{W}_{\text{B}} \) is initialized with zeros. By initializing \( \mathcal{W}_A \) with SVD decomposition, CABR-LoRA effectively preserves prior knowledge. The initialization of \( \mathcal{W}_B \) with zeros ensures that this matrix will fine-tune from the base LLM without altering the critical knowledge.
The decomposition follows the formulas below:
\begin{equation}
    \mathcal{W}_{\text{frozen}} = \mathcal{U}_{\text{SVD}} \mathcal{S}_{\text{SVD}} \mathcal{V}_{\text{SVD}}^{\top}.
\end{equation}
\begin{equation}
    \mathcal{W}_A = \mathcal{U}_{\text{SVD}}^{(r,:)} \mathcal{S}_{\text{SVD}} \mathcal{V}_{\text{SVD}}^{(:,m) \top}.
\end{equation}
\begin{equation}
    \mathcal{W}_B = \mathbf{0}_{m \times r}, \quad \text{where} \quad \mathbf{0}_{m \times r} \text{ is a zero matrix}.
\end{equation}

\subsection{S-MagNorm Normalization Algorithm}

While CABR decomposition provides a higher dimension of fine-tuning,  only using CABR-LoRA alone does not fully address catastrophic forgetting, which is crucial for performance in lifelong learning. To tackle this, we introduce the S-MagNorm Normalization algorithm, inspired by dynamic network pruning. This algorithm adjusts the weight updates by controlling the change magnitude between the former and new weights, provide a stable parameter update (figure~\ref{fig:Grad_analysis}).

The core idea is to employ a Sigmoid-based dynamic scaling mechanism to regulate weight updates. This method constructs a weight pruning matrix whose values are constrained within the range \([1,2]\) to achieve the task of weighted mean between basic matrix weight and updated matrix weight. The Sigmoid function primarily emphasizes central value adjustments, effectively suppressing extreme variations. We can allow SECURA deal with whole matrix's changing tendency while limiting the extreme parameters' effect. Simultaneously, restricting the update amplitude of lower-norm columns and rows ensures that important parameters(with higher norms)'  magnitude experience minimal change, while less important parameters (with smaller norms) undergo more significant updates. The base weight matrix is denoted as \( W_{\text{base}} \), and the SECURA's matrices as \( A \) and \( B \).

The merged weight matrix is computed as:
\begin{equation}
    \mathcal{W}_{\text{SECURAMerged}} = CABR + \mathcal{W}_{\text{base}}.
\end{equation}
Next, we calculate the Magnitude Loss Matrix \(\mathcal{M}_{\text{Mag}}\), which captures the relative magnitude change between the new and base weights. Here \(\epsilon\) is a  small value to avoid division by too small parameters:
\begin{equation}
    \mathcal{M}_{\text{Mag}} = \left| \frac{\mathcal{W}_{\text{SECURAMerged}}}{\mathcal{W}_{\text{base}} + \epsilon} \right|.
\end{equation}
Then, we normalize the \(\mathcal{M}_{\text{Mag}}\) to the range \([-0.5, 0.5]\) to balance the values on both sides of zero:
\begin{equation}
    \mathcal{M}_{\text{norm}} = \left[ \frac{\mathcal{M}_{\text{Mag}}}{\max(\mathcal{M}_{\text{Mag}}) + \epsilon} - 0.5 \right] \cdot \text{Scale}.
\end{equation}
Note that the Scale parameter here resizes the normed value to ensure it falls within the range \([0,1]\) after applying the Sigmoid function, as the Sigmoid function only produces a range of \([0.3775, 0.6225]\) for the \([-0.5, 0.5]\) interval.

Finally, we apply the Sigmoid function to the normalized matrix to produce the S-MagNorm values and get the final restring matrix \(\mathcal{M}_{\text{Res}}\):
\begin{equation}
    \mathcal{M}_{\text{Res}} = 2 - \sigma\left(\mathcal{M}_{\text{norm}}\right).
\end{equation}

Once we have weight pruning matrix \(\mathcal{M}_{\text{Res}}\) , the mergred matrix \( W_{\text{SECURAMerged}} \), will then adjust the update process by dividing the S-MagNorm pruning-weight matrix \(\mathcal{M}_{\text{Res}}\):
\begin{equation}
    \mathcal{W}_{\text{updated}} = \frac{\mathcal{W}_{\text{SECURAMerged}}}{\mathcal{M}_{\text{Res}}}.
\end{equation}
This dynamic update process allows controlled pruning of weights, where updates for important parameters are slowed down and less important one remains, helping preserve important knowledge and mitigating catastrophic forgetting(Figure ~\ref{fig:SVD_Comparison}).

\subsection{Merge with Basic Model}
Limiting parameter growth alone is not optimal. Our early experiments showed that using only CABR-LoRA and S-MagNorm degraded both base performance and catastrophic forgetting mitigation. Drawing insights from delayed updates~\citep{kosta2020cost} and Target Network~\citep{gao2017adaptive}, we recognized that base parameters require controlled updates to prevent performance stagnation.

When freezing base parameters, a destructive two-phase cycle emerges: 
1) Unimportant parameters update freely while important ones remain constrained. 
2) Important parameters become smaller than unimportant ones in subsequent cycles, accelerating catastrophic forgetting as critical knowledge degrades.

We resolve this through two merging strategies:
\paragraph{M1: Direct Weight Merge}
Updates base LLM weights with one SECURA training cost:
\begin{equation}
\mathcal{W}_{\text{updated}} = \underbrace{C A_{\text{former}} B_{\text{train}} R}_{\text{CABR-LoRA}} + \mathcal{W}_{\text{base}}
\end{equation}

\paragraph{M2: Frozen Base Merge} 
Preserves base weights using dual SECURA instances: 
\begin{equation}
\begin{split}
\mathcal{W}_{\text{merged}} =& \underbrace{C A_{\text{frozen}} B_{\text{accum}} R}_{\text{Previous CABR-LoRA}} + \underbrace{C A_{\text{train}} B_{\text{reset}} R}_{\text{New CABR-LoRA}} \\
&\hspace{11em} + \mathcal{W}_{\text{base}}
\end{split}
\end{equation}
where parameter updates follow:
\begin{align}
A_{\text{former}},A_{\text{frozen}} & \leftarrow A_{\text{train}} \\
B_{\text{accum}}                    & \leftarrow B_{\text{accum}} + B_{\text{train}} \\
B_{\text{train}}                    & \leftarrow \mathbf{0} 
                                      \quad \text{\footnotesize (Post-update reset)}
\end{align}
M1 enables stronger fine-tuning by modifying base weights, while M2 prioritizes knowledge retention through:
\begin{compactitem}
\item \textbf{Parameter isolation}: Base weights remain frozen
\item \textbf{Progressive accumulation}: $B_{\text{accum}}$ aggregates historical updates
\item \textbf{Cyclic resetting}: $B_{\text{train}}$ reinitialization prevents update saturation
\end{compactitem}

Our ablation study (\ref{tab:Merge_method_Ablation}) reveals task-dependent superiority:
\begin{compactitem}
\item \textbf{M1} for performance-critical domains (mathematics, programming)
\item \textbf{M2} for knowledge-sensitive tasks (medical, economics) 
\end{compactitem}
The hybrid approach balances both objectives, with M1 generally providing better balanced optimization results.

%% file: sections/experiment.tex
\begin{figure}[tb]
    \centering
    \includegraphics[width=\columnwidth]{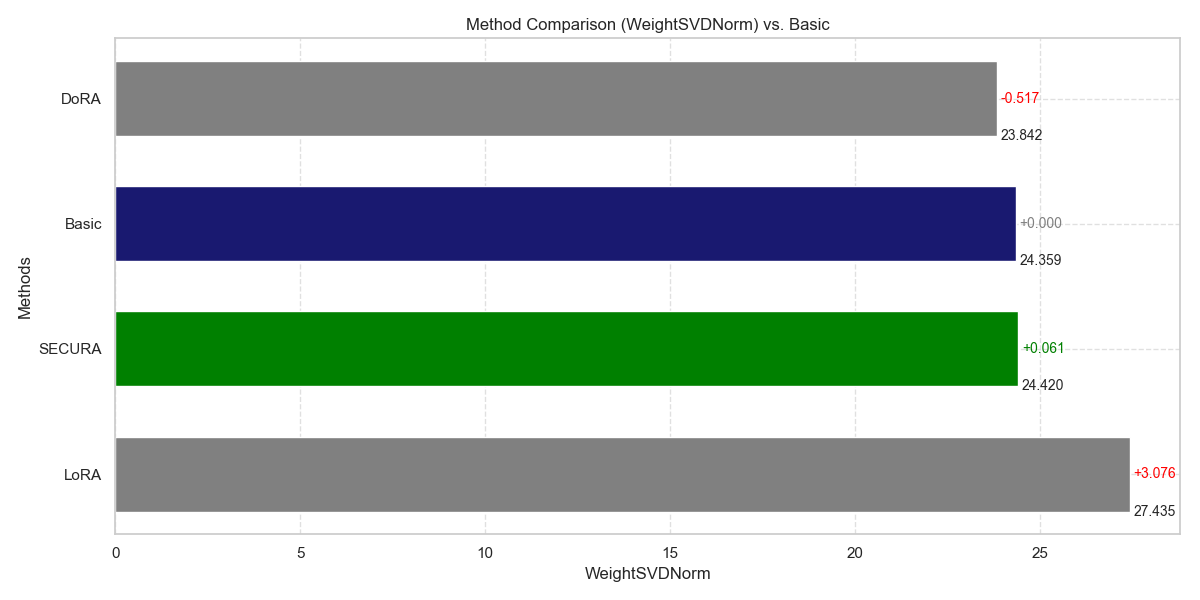}
    \caption{Comparison of Weight SVD Norm across different methods: The "Basic" method represents the original LLM weights without fine-tuning. After 2000 training steps, DoRA shows a 0.517 decrease in norm, indicating some loss of information. SECURA preserves more knowledge with a minor 0.061 change, while LoRA shows a significant 3.076 increase, reflecting greater weight modification. Hyperparameter settings are detailed in Appendix \ref{app:hyper_param}.}
    \label{fig:SVD_Comparison}
    \vspace{-1em}
\end{figure}

\section{Experiments}

\begin{figure*}[tb]
    \centering
    \includegraphics[width=\textwidth]{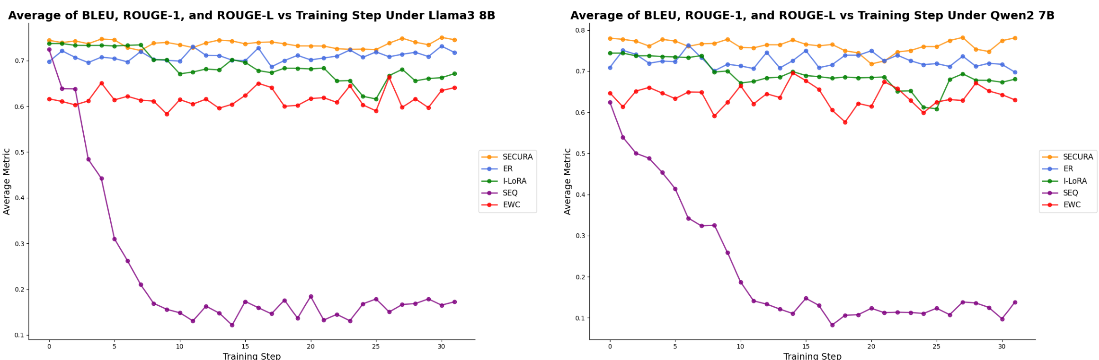}
    \caption{The performance of SECURA compared to  16 tasks baselines sequentially trained using LLaMA-3 8B (left) and Qwen-2 7B (right) backbones, tested under F-task with learning rate 1e-5. Detailed results are in Appendix~\ref{app:all_results}}
    \label{fig:EXP3}
\end{figure*}

\subsection{Experimental Setup}

\paragraph{Datasets.}

We conducted comprehensive evaluations on 18 datasets: 15 from BBM~\citep{srivastava2023beyond},  while remaining datasets are selected from Llama3's~\citep{dubey2024llama}, Qwen2's~\citep{qwen2.5}, Gemma2's~\citep{team2024gemma} base training or fine-tuning datasets.  These datasets include 8 multiple-choice question (MCQ) datasets and 10 question-answering (QA) datasets,  spanning six domains: mathematical reasoning(GSM8K)~\citep{cobbe2021training},  logical deduction(LogicalQA)~\citep{srivastava2023beyond},  multilingual identification(CNN/DailyMail)~\citep{see2017get},  programming challenges(Multipl-E)~\citep{cassano2022multiplescalableextensibleapproach},  translation questions(News Commentary dataset)~\citep{tiedemann2012parallel},  and commonsense understanding(CommonsenseQA)~\citep{talmor-etal-2019-commonsenseqa}. Specifically, all of our benchmarks are carefully handled, with 1000 samples taken from each respective source and a 9:1 training-to-testing split.

To evaluate catastrophic forgetting, we combined three QA datasets-ARC-Challenge(ARC-C)~\citep{clark2018think}, GSM8K~\citep{cobbe2021training}, Multipl-E~\citep{cassano2022multiplescalableextensibleapproach}-as the test set , which were used for training/fine-tuning the selected LLMs. This fusion dataset will be referred to as  \textbf{F-task} in following parts. Further details can be found in the appendix~\ref{app:18_datasets}. 

\paragraph{LLM Backbones, LoRA variants, and Knowledge retention method.}

Our study employs five state-of-art LLMs: Gemma2-2B, Qwen2-1.5B/7B, and Llama3-8B/3.1-8B.

We evaluate our SECURA framework using four PEFT variants: 

\begin{compactitem}
    \item LoRA~\citep {whitehouse-etal-2024-low}: Standard low-rank adaptation.
    \item DoRA~\citep{liu2024doraweightdecomposedlowrankadaptation}: Weight-Decomposed Low-Rank Adaptation.
    \item CUR-LoRA~\citep{https://doi.org/10.5281/zenodo.12730055}: CUR-Decomposed low-rank adaptation.
    \item I-LoRA~\citep{ren2024analyzingreducingcatastrophicforgetting}: Interpolation-based low-rank adaptation.
\end{compactitem}

Additionally, we implement five continuous learning baselines:
\begin{compactitem}
\item SEQ~\citep{sutskever2014sequencesequencelearningneural}:Sequential Learning
\item EWC~\citep{aich2021elasticweightconsolidationewc}:Elastic Weight Consolidation 
\item ER~\citep{fedus2020revisitingfundamentalsexperiencereplay}:Experience Retention
\item CUR-LoRA~\citep{https://doi.org/10.5281/zenodo.12730055}:CUR-Decomposed low-rank adaptation
\item I-LoRA~\citep{ren2024analyzingreducingcatastrophicforgetting}:Interpolation-based low-rank adaptation.
\end{compactitem}

\paragraph{Evaluation Metrics and Main Experiments}

We design three main experiments to assess SECURA's performance:

\textbf{Experiment 1 (PEFT Comparison):} Compares LoRA variants through 5 basic model backbones with 4 MCQ measured by accuracy (AC) and 5 QA performance measured by BLEU, ROUGE-1, ROUGE-L.

\textbf{Experiment 2 (Catastrophic Forgetting with PEFT Comparison):} In order to clearly evaluates knowledge retention with unified training extreme environment  (learning rate 1e-3)  under different PEFT method, we use the F-task set measured by BLEU, ROUGE-1, ROUGE-L.

\textbf{Experiment 3 (Catastrophic Forgetting with Knowledge retention method):} Benchmarks against continual learning methods at 1e-5 learning rate with sequential training on 16 tasks (each trained for 2K steps, totaling 32K steps) every PEFT based training method will fusion itself to frozen basic weights(e.g. CUR-LoRA, I-LoRA, SECURA) after one training task is done, measuring remain knowledge by using the F-task set and measured by BLEU, ROUGE-1, ROUGE-L.

Note: All main experiments' SECURA uses the M1 merge method. For hyperparameter details, refer to the appendix~\ref{app:hyper_param}.

\subsection{Main Result}

\begin{table}[tb]
    \centering
    \small
    \resizebox{\columnwidth}{!}{
    \begin{tabular}{lccccc}
    \toprule
    \textbf{Task} & \textbf{LoRA} & \textbf{DoRA} & \textbf{CUR-LoRA} & \textbf{ILoRA} & \textbf{SECURA} \\
    \midrule
    \multicolumn{6}{c}{\textbf{MCQ Avg.}} \\
    AbsNarr    
& 83.81 & 85.00 & 64.06 & 80.70 & \textbf{87.94} 
\\
    ObjCount   
& 86.49 & 86.43 & 59.35 & 82.49 & \textbf{89.40} 
\\
    PlayDiag   
& 85.71 & 84.93 & 56.53 & 76.74 & \textbf{89.87} 
\\
    ColorReason & 80.44 & 80.86 & 58.30 & 79.64 & \textbf{84.37}\\
    \bottomrule
    \multicolumn{6}{c}{\textbf{QA Avg.}} \\
    GSM8K & 81.11& 81.53& 66.79
& 38.21
& \textbf{82.11}\\
    NewsDE & 62.06
& 61.81
& 60.68
& 40.60
& \textbf{64.20}
\\
    NewsIT & 65.19
& 65.78
& 63.75
& 42.66
& \textbf{66.43}
\\
    NewsES & 60.25
& 58.54
& 60.70
& 39.47
& \textbf{66.06}
\\
    ALPACA & 53.22& 54.41& 41.09
& 26.53
& \textbf{55.82}
\\
    \bottomrule
    \end{tabular}}
   \caption{Performance comparison of PEFT variants under 5 LLMs across 4 MCQ tasks and 5 QA tasks. MCQ shows accuracy, QA shows BLEU, ROUGE-1, and ROUGE-L averages across all 5 LLMs. Detailed results are in Appendix~\ref{app:all_results}.}
    \label{tab:EXP1_AVG}
\end{table}

\paragraph{Performance on PEFT Comparison}

As shown in Table~\ref{tab:EXP1_AVG} SECURA outperforms all PEFT methods across both MCQ and QA tasks, achieving SOTA results. Compared to basic lora training, SECURA get average improvement 3.63\% in MCQ tasks and gains 2.56\% improvement in QA tasks. Notably, on tasks of NewsES, SECURA demonstrates nearly a 6\% improvement,  highlighting its exceptional fine-tuning capabilities. In contrast, other variants which major in handle the problem of catastrophic forgetting got negative improvement, even those focused on fine-tuning performance like DoRA, shows minor improvements, which only gains most improvement 1.19\% in abstract narrative understanding.

\begin{table}[tb]
    \centering
    \small
    \resizebox{\columnwidth}{!}{
    \begin{tabular}{lccccc}
    \toprule
    \textbf{Task} & \textbf{LoRA} & \textbf{DoRA} & \textbf{CUR-LoRA} & \textbf{ILoRA} & \textbf{SECURA} \\
    \midrule

    AbsNarr
& 00.80
& 01.19& 04.48& 09.33& \textbf{72.09}
\\
    DisflQA
& 06.59
& 23.93
& 10.54
& 26.80
& \textbf{72.34}
\\
    LangID
& 00.47
& 03.95& 08.29& 24.32
& \textbf{72.42}
\\
    LogDeduc
& 01.03
& 00.30& 19.01
& 25.82
& \textbf{72.66}
\\
    ObjCount
& 08.10
& 07.88& 21.44
& 42.59
& \textbf{71.18}
\\
    PlayDiag
& 00.22
& 00.71& 05.97& 33.96
& \textbf{72.43}
\\
    ColorReason
& 01.17
& 00.28& 08.56& 39.18
& \textbf{72.31}
\\
    TrackObj& 00.26& 01.53& 03.22& 21.77
& \textbf{72.26}
\\
    \bottomrule
    \end{tabular}}
   \caption{Knowledge retention comparison with PEFT variants under 5 LLMs, using 8 tasks separated trained and tested on the F-task. Performance is measured by BLEU, ROUGE-1, and ROUGE-L averages across all 5 LLMs. Detailed results are in Appendix~\ref{app:all_results}.}
    \label{tab:EXP2 AVG}
\end{table}

\paragraph{Catastrophic Forgetting with PEFT Comparison}

To clearly shows SECURA's capability of Knowledge retention, we create a extreme environment(learning rate is 1e-3) through paper~\citep{Kirkpatrick_2017},~\citep{8107520},~\citep{zenke2017continuallearningsynapticintelligence} which shows with higher learning rate, the risk of catastrophic raised. The performance shows in the table~\ref{tab:EXP2 AVG}.

As shown in table, most of PEFT variants almost lose all of its former knowledge, even the variants which aimed to alleviate catastrophic forgetting such as CUR-LoRA and I-LoRA are still have heavily forgetting and only can keep average score under 42.59\%, while SECURA shows 
strong capabilities of limit catastrophic forgetting even in such tense environment and gains average score of 72.21\% under the test of F-task. Demonstrating its robustness in mitigating catastrophic forgetting.

\paragraph{Catastrophic Forgetting with Knowledge retention method}

SECURA is compared with other knowledge retention methods across 16 sequential training tasks, tested under the F-task. As shown in Figure~\ref{fig:EXP3}, SECURA consistently outperforms other methods in retaining knowledge, with only slight improvements over ER and I-LoRA. However, SECURA maintains superior knowledge retention and fine-tuning capability, making it an effective method for fine-tuning LLMs while limiting catastrophic forgetting.

\subsection{Ablation Study}
We have already conducted the ablation study comparing between Gradient variations   and SVD norm in figure ~\ref{fig:Grad_analysis} and ~\ref{fig:SVD_Comparison}. Here, we provide a more detailed ablation study focusing on the issues arising from different merge methods. 

\paragraph{How to Prove the merge hypothesis of the Update Cycle?}

\begin{table}[tb]
    \centering
    \small
    \resizebox{\columnwidth}{!}{
    \begin{tabular}{lcc|cc}
    \toprule
    \textbf{Task} & \textbf{SECURA-1} & \textbf{SECURA-200} & \textbf{SECURA-1} & \textbf{SECURA-200} \\
    \midrule
    \multicolumn{3}{c}{\textbf{Qwen-2 7B}} & \multicolumn{2}{c}{\textbf{Llama-3.1 8b}} \\
            AbsNarr&  \textbf{77.93}
& 07.36&  \textbf{74.73}
& 54.17
\\
    DisflQA&  \textbf{78.46}
& 70.98
&  \textbf{73.85}
& 66.99
\\
    LangID&  \textbf{78.86}
& 00.16&  \textbf{73.13}
& 4.866
\\
    LogDeduc&  \textbf{75.19}
& 54.61
&  \textbf{75.11}
& 54.46
\\
    ObjCount&  \textbf{78.30}
& 73.30
&  \textbf{73.41}
& 61.19
\\
    PlayDiag&  \textbf{78.10}
& 00.22&  \textbf{74.81}
& 62.30
\\
    ColorReason&  \textbf{77.81}
& 00.26&  \textbf{75.10}
& 39.13
\\
    TrackObj&  \textbf{78.38}
& 74.29
&  \textbf{74.35}
& 65.78
\\
    \bottomrule
    \end{tabular}}
    \caption{Ablation study comparing knowledge retention performance of 1-step and 200-step merges under EXP2. The 1-step merge achieves better performance with minimal forgetting, while the 200-step merge leads to a cycle with reduced performance}
    \label{tab:Ablation_Cycle}
\end{table}

To prove the cycle hypothesis, we conducted an experiment comparing the performance of 1-step and 200-step merges across different LLMs. The results are shown in the table~\ref{tab:Ablation_Cycle}. Although the 200-step merge still retains some capabilities, it has obviously lower performance compared to the 1-step merge. We hypothesize that this performance degradation occurs because in the 200-step merge, the system has started entering the second cycle but not fully tracked by the problem. This suggests that the merge function plays a critical role in preventing catastrophic forgetting by maintaining a stable learning cycle.

\paragraph{Different Performance based on different Merge method}
\begin{table}[tb]
    \centering
    \small
    \resizebox{\columnwidth}{!}{
    \begin{tabular}{lcc}
    \toprule

    \textbf{Merge Under EXP1:} & \textbf{MergeMethod-1} & \textbf{MergeMethod-2} \\
    \midrule
    AbsNarr    
& \textbf{87.48}
& 77.27\\
    ObjCount   
& \textbf{85.21}
& 79.83\\
    PlayDiag   
& \textbf{90.03}
& 74.00
\\
    ColorReason & \textbf{88.53} & 70.37\\
    \midrule
    \textbf{Merge Under EXP2:} & \textbf{MergeMethod-1} & \textbf{MergeMethod-2} \\
    \midrule
    AbsNarr     & 64.18 / 63.77 / 50.89
& \textbf{74.47 / 74.71 / 69.06}\\
    ObjCount    & 62.37 / 63.15 / 50.86
& \textbf{76.12 / 76.01 / 70.01}
\\
    PlayDiag    & 63.56 / 64.47 / 51.53
& \textbf{75.15 / 75.32 / 69.32}
\\
    ColorReason & 64.93 / 64.64 / 51.72
& \textbf{75.91 / 75.90 / 69.86}\\

    \bottomrule
    \end{tabular}}
    \caption{Comparison of fine-tuning efficiency and knowledge retention using MergeMethod-1 and MergeMethod-2 under two experiments (EXP1 and EXP2).}
    \label{tab:Merge_method_Ablation}
\end{table}
We designed another ablation study by testing different performance under M1 and M2 merge method with Gemma2 2b backbone and shows its feedback on table~\ref{tab:Merge_method_Ablation}. 

As the prediction in methodology, M1 method with deepen merge which will change basic weights shows more benefits on fine-tuning, performance better than M2 method range from 5.62\% to 18.16\%. In contrast M2 method which with shallow merge method shows more advantage under knowledge retention capabilities. Where it average improved retention 11.65\% in BLEU, 11.48\% in ROUGE-1 and 18.31\% in ROUGE-L  compared to M1 method.

%% file: sections/discussion.tex
\section{Discussion}

\subsection{Why Do We Update the Basic Parameters or Track SECURA Parameters But Never Change \(C\) and \(R\)?}

We assume that parameters with higher norms carry more important information while lower-norm parameters are considered less important. The columns and rows \( C \) and \( R \) we initialized are represent the less important parameters' positions. Our goal is to keep the former important knowledge intact, which means it is necessary to prevent changes to \( C \) and \( R \) which are used to identify the locations that should be updated. By keeping \( C \) and \( R \) unchanged, we ensure that the critical information in the LLM is protected, while non-critical parameters can be updated or fine-tuned as needed.

\subsection{What Are the Other Benefits of SECURA Compared to Other Catastrophic Forgetting Methods?}

Since we are already shows SECURA's performance in our experiments. Here, we will point out the benefits of using SECURA over other methods like EWC or Experience Replay, specifically in terms of efficiency and simplicity.

\subsubsection{Smaller Weight}

Traditional methods like EWC rely on the Fisher matrix to detect changes between current and previous parameters, requiring the storage of previous parameters, which increases computational costs and space requirements. In contrast, SECURA as a PEFT fine-tuning method can be outperforming with even lower parameters compared to basic LoRA training~\ref{app:hyper_param}. 

\subsubsection{No Need for Former Datasets and Calculation}

Methods like Experience Replay store and reuse previous datasets to prevent catastrophic forgetting, but this comes at the cost of time and computational resources. SECURA, however, does not require past data and still performs effectively. This eliminates the need for dataset storage and reuse, making SECURA a more resource-efficient solution.



%% file: sections/appendix.tex
\section{Broader Impacts}
\begin{itemize}
    \item \textbf{Efficiency and Accessibility:} By eliminating the need for prior datasets and reducing training time for knowledge retention, SECURA lowers the computational barrier for fine-tuning LLMs, making advanced AI adaptation more accessible to researchers with limited resources.
    
    \item \textbf{Environmental Sustainability:} The reduced computational overhead  compared to Experience Replay could decrease energy consumption in large-scale LLM deployments.
    
    \item \textbf{Ethical Risks:} While SECURA mitigates catastrophic forgetting, its ability to retain sensitive information from pre-training data (e.g., biased or private content) requires careful auditing before deployment. Future work should investigate selective forgetting mechanisms to align with ethical AI principles.
    
    \item \textbf{Generalization Beyond NLP:} The S-MagNorm algorithm, though initially designed for PEFT, could be extended to other continual learning scenarios (e.g., robotics, healthcare), potentially improving adaptive systems in dynamic environments.
\end{itemize}

\input{sections/secura_algorithm.tex}
\section{Pseudo code deduction}\label{app:Pseudo}
Algorithm~\ref{alg:secura} shows whole deduction of SECURA:

\section{Details about 18 Tasks and Datasets}\label{app:18_datasets}
Table~\ref{tab:18_tasks_details} includes detailed descriptions of each dataset's name, keywords, main content and corresponding evaluation metrics. These 18 tasks include different topics, such as mathematical reasoning,  logical deduction,  multilingual identification,  programming challenges,  translation questions,  and commonsense understanding.

\begin{table*}[!h]
\centering
\renewcommand{\arraystretch}{1.4}
\resizebox{\textwidth}{!}{
\begin{tabular}{cp{3.5cm}p{7cm}p{2.8cm}}
\toprule
Task Name & Keywords & Description & Evaluation Metrics \\ 
\midrule
abstract\_narrative\_understanding (AbsNarr) &
  narrative understanding, multiple choice &
  Given a narrative, choose the most related proverb. &
  Accuracy \\
cnn\_dailymail (CNNDM) &
  summarization &
  Given news articles, write the summarization. &
  ROUGE \\
disfl\_qa (DisflQA) &
  contextual question-answering, reading comprehension &
  Pick the correct answer span from the context given the disfluent question. &
  Accuracy \\
gsm8k (GSM8K) &
  mathematics &
  Solve the grade school math word problems. &
  Accuracy \\
language\_identification (LangID) &
  multilingual, multiple choice &
  Given a sentence, select the correct language. &
  Accuracy \\
linguistics\_puzzles (LingPuzz) &
  logical reasoning, linguistics &
  Solve Rosetta Stone-style linguistics puzzles. &
  BLEU, ROUGE \\
logical\_deduction (LogDeduc) &
  logical reasoning, multiple choice &
  Deduce the order of a sequence of objects. &
  Accuracy \\
news\_commentary\_de (NewsDE) &
  multilingual, translation &
  Translate German sentences into English. &
  BLEU \\
news\_commentary\_es (NewsES) &
  multilingual, translation &
  Translate Spanish sentences into English. &
  BLEU \\
news\_commentary\_it (NewsIT) &
  multilingual, translation &
  Translate Italian sentences into English. &
  BLEU \\
object\_counting (ObjCount) &
  logical reasoning &
  Questions that involve enumerating objects and asking the model to count them. &
  Accuracy \\
play\_dialog\_same\_or\_different (PlayDiag) &
  reading comprehension, multiple choice &
  Determine if nearby lines in a Shakespeare play were spoken by the same individual. &
  Accuracy \\
question\_selection (QuestSel) &
  reading comprehension, multiple choice &
  Given an answer along with its context, select the most appropriate question which has the given answer as its answer. &
  Accuracy \\
reasoning\_about\_colored\_objects (ColorReason) &
  reading comprehension, logical reasoning, multiple choice &
  Answer extremely simple questions about the colors of objects on a surface. &
  Accuracy \\
strategyqa (StratQA) &
  logical reasoning, context-free question answering &
  Answer questions in which the required reasoning steps are implicit in the question. &
  BLEU, ROUGE, Accuracy \\
tracking\_shuffled\_objects (TrackObj) &
  logical reasoning, multiple choice &
  Determine the final positions given initial positions and a description of a sequence of swaps. &
  Accuracy \\
Abstraction and Reasoning Corpus - Challenge (ARC-C) &
  logical reasoning, multiple choice &
  Answer questions that require reasoning about complex, abstract scenarios, often requiring multi-step logic and understanding of concepts. &
  Accuracy \\
Multiple-Choice Evaluations for Embodied Agents (Multiple-E) &
  reasoning, multiple choice, dynamic problems &
  Answer questions based on a sequence of logical events or transformations, with an emphasis on evolving scenarios and inference. &
  Accuracy \\
\bottomrule
\end{tabular}}
\caption{Details about the 18 selected tasks following~\cite{zhang2025dlploraefficienttaskspecificlora}.}
\label{tab:18_tasks_details}
\end{table*}

\section{Experimental hyper parameters on three main experiments}\label{app:hyper_param}
All of our main experiments conducted on two NVIDIA 2080 Ti 11GB GPUs, table  ~\ref{tab:hyper parameters_1}, ~\ref{tab:hyper_parameters_2}, ~\ref{tab:hyper_parameters_3} and  ~\ref{tab:hyper_parameters_4} show all hyper-parameters among different LoRA baselines and SECURA using in experiments.

\begin{table*}[tb]
    \centering
    \small
    \resizebox{\textwidth}{!}{
    \begin{tabular}{lccccc}
    \toprule
    Hyper
Parameters& \textbf{LoRA} & \textbf{DoRA} & \textbf{CUR-LoRA} & \textbf{ILoRA} & \textbf{SECURA} \\
    \midrule
    \multicolumn{6}{c}{Exp 1} \\
    \bottomrule
    Rank& 8& 8& -& 8& -\\
    learning rates& 1e-5& 1e-4& 1e-4& 1e-3& 1e-3\\
    CABR/CUR r Rank& -& -& 150& -& 150\\
 CABR m Rank& -& -& -& -&200\\
 Fusion Step& -& -& -& -&1\\
 Trainable parameters(each layer)& 1.31e+5& 1.48e+5& 0.90e+5& 2.62e+5&1.20e+5\\
    \bottomrule
    \multicolumn{6}{c}{Exp 2} \\
    \bottomrule
    Rank
& 8& 8& -& 8& -
\\
    learning rates
& 1e-3& 1e-3& 1e-3& 1e-3& 1e-3
\\
    CABR/CUR r Rank& -& -& 150& -& 150
\\
 CABR m Rank& -& -& -& -&200\\
 Fusion Step& -& -& -& -&1\\
 Trainable parameters(each layer)& 1.31e+5& 1.48e+5& 0.90e+5& 2.62e+5&1.20e+5\\
    \bottomrule
    \end{tabular}}
   \caption{hyper-parameters of experiment 1\&2}
    \label{tab:hyper parameters_1}
\end{table*}

\begin{table*}[tb]
    \centering
    \small
    \resizebox{\textwidth}{!}{
    \begin{tabular}{lcccccc}
    \toprule
    \textbf{Exp 3} & \textbf{SEQ} & \textbf{EWC} & \textbf{ER} & \textbf{ILoRA} & \textbf{CUR-LoRA} & \textbf{SECURA} \\
    \midrule
    Rank                                & -& -& -& 8     & -     & -     \\
    Learning Rates                      & 1e-5  & 1e-5  & 1e-5  & 1e-5  & 1e-5  & 1e-5  \\
    CABR/CUR r Rank                     & -     & -     & -& -     & 150   & 150   
\\
    CABR m Rank                          & -     & -     & -     & -     & 200   & 200   \\
    Fusion Step                          & 2000  & 2000  & 2000  & 1& 2000& 1     \\
    Trainable Parameters (each layer)    & FT& FT& FT& 2.62e+5 & 1.20e+5 & 1.00e+5 \\
    \bottomrule
    \end{tabular}}
    \caption{Hyper-parameters of experiment 3}
    \label{tab:hyper_parameters_2}
\end{table*}

\begin{table*}[tb]
    \centering
    \small
    \resizebox{\textwidth}{!}{
    \begin{tabular}{lccc}
    \toprule
    Gradient Exp& LoRA& CABR-LoRA& \textbf{SECURA} \\
    \midrule
    Rank                                & 8     & -     & -     \\
    Learning Rates                      & 1e-3& 1e-3& 1e-3\\
    CABR/CUR r Rank                     & -     & 150   & 150   \\
    CABR m Rank                          & -     & 200   & 200   \\
    Fusion Step                          & -& -& 1     \\
    Trainable Parameters (each layer)    & 2.62e+5 & 1.20e+5 & 1.00e+5 \\
    \bottomrule
    \end{tabular}}
    \caption{Hyper-parameters of Gradient comparison experiments}
    \label{tab:hyper_parameters_3}
\end{table*}

\begin{table*}[tb]
    \centering
    \small
    \resizebox{\textwidth}{!}{
    \begin{tabular}{lcccc}
    \toprule
    SVD Norm Exp& LoRA &DoRA& CABR-LoRA& \textbf{SECURA} \\
    \midrule
    Rank                                & 8      & 8      & -     & -     \\
    Learning Rates                      & 1e-3 & 1e-3 & 1e-3& 1e-3\\
    CABR/CUR r Rank                     & -      & - & 150   & 150   \\
    CABR m Rank                          & -      & - & 200   & 200   \\
    Fusion Step                          & - & - & -& 1     \\
    Trainable Parameters (each layer)    & 2.62e+5  & 1.48e+5 & 1.20e+5 & 1.00e+5 \\
    \bottomrule
    \end{tabular}}
    \caption{Hyper-parameters of SVD Norm experiments}
    \label{tab:hyper_parameters_4}
\end{table*}

\section{Experimental Results on All Datasets}\label{app:all_results}
Table~\ref{tab:EXP1-MCQ},~\ref{tab:EXP1-QA},~\ref{tab:EXP2-Forgetting}, ~\ref{tab:EXP3 Llama3 8B} and ~\ref{tab:EXP3 Qwen-2 7B} show all results among different LoRA baselines and SECURA using Gemma2 2b, Qwen2 1.5B, Qwen2 7B, LLaMA3 7B and LLaMA3.1 8B backbones.

\begin{table*}[tb]
    \centering
    \small
    \resizebox{\textwidth}{!}{
    \begin{tabular}{lccccc|ccccc}
    \toprule
    \textbf{Task} & \textbf{LoRA} & \textbf{DoRA} & \textbf{CUR-LoRA} & \textbf{ILoRA} & \textbf{SECURA} & \textbf{LoRA} & \textbf{DoRA} & \textbf{CUR-LoRA} & \textbf{ILoRA} & \textbf{SECURA}\\
    \midrule
    \multicolumn{6}{c}{\textbf{Gemma-2 2B}} & \multicolumn{5}{c}{\textbf{Llama-3 8B}}\\
            AbsNarr    & 68.67 & 72.43 & 37.03 & 73.57 & 87.48 & 90.35 & 90.91 & 71.94 & 90.06 & 89.54 \\
            ObjCount   & 78.74 & 77.34 & 58.34 & 78.64 & 85.21 & 89.43 & 89.76 & 74.08 & 89.52 & 92.83\\
            PlayDiag   & 72.40 & 72.77 & 53.39 & 37.57 & 90.03 & 91.77 & 90.26 & 73.38 & 88.65 & 92.49\\
            ColorReason & 71.00 & 70.93 & 39.27 & 69.90 & 88.53 & 88.07 & 87.11 & 71.43 & 87.09 & 88.60 \\\midrule
    \multicolumn{6}{c}{\textbf{Qwen-2 1.5B}} & \multicolumn{5}{c}{\textbf{Llama-3.1 8B}}\\
            AbsNarr    & 84.07 & 85.67 & 34.13 & 77.13 & 86.72 & 89.43 & 88.95 & 76.81 & 89.23 & 88.65\\
            ObjCount   & 84.38 & 85.72 & 36.24 & 73.57 & 85.01 & 89.51 & 89.68 & 74.23 & 89.55 & 93.44 \\
            PlayDiag   & 85.20 & 84.40 & 40.27 & 83.87 & 85.72 & 91.51 & 91.97 & 71.60 & 89.94 & 92.51 \\
            ColorReason & 74.60 & 76.47 & 52.60 & 77.22 & 75.38 & 88.12 & 87.46 & 72.07 & 87.14 & 88.34 \\\midrule
    \multicolumn{6}{c}{\textbf{Qwen-2 7B}} & \multicolumn{5}{c}{\textbf{Avg.}} \\
            AbsNarr    & 86.53 & 87.06 & 40.41 & 73.49 & 87.33 & 83.81 & 85.00 & 64.06 & 80.70 & 87.94 \\
            ObjCount   & 90.38 & 89.67 & 53.85 & 81.16 & 90.50 & 86.49 & 86.43 & 59.35 & 82.49 & 89.40 \\
            PlayDiag   & 87.67 & 85.27 & 44.00 & 83.65 & 88.62 & 85.71 & 84.93 & 56.53 & 76.74 & 89.87\\
            ColorReason & 80.40 & 82.33 & 56.13 & 76.83 & 81.00 & 80.44 & 80.86 & 58.30 & 79.64 & 84.37\\
    \bottomrule
    \end{tabular}}
    \caption{EXP1-MCQ}
    \label{tab:EXP1-MCQ}
\end{table*}

\begin{table*}[tb]
    \centering
    \small
    \resizebox{\textwidth}{!}{
    \begin{tabular}{lccccc}
    \toprule
    \textbf{Task} & \textbf{LoRA} & \textbf{DoRA} & \textbf{CUR-LoRA} & \textbf{ILoRA} & \textbf{SECURA} \\
    \midrule
    \multicolumn{6}{c}{\textbf{Gemma-2 2B}} \\
    GSM8K & 58.38 / 56.01  / 50.74& 60.88 / 58.20  / 53.73 & 57.03 / 55.28  / 49.41& 60.22 / 58.76  / 53.86& 58.84 / 56.25  / 50.95
\\
    NewsDE & 63.42 / 62.57  / 57.20& 63.53 / 62.53  / 56.74& 64.02 / 63.84  / 57.81& 62.82 / 62.33  / 56.69& 61.98 / 60.69  / 51.41
\\
    NewsIT & 65.51 / 64.25  / 62.96& 66.02 / 65.71  / 64.50& 67.73 / 66.26  / 65.16& 65.06 / 63.68  / 62.25& 61.94 / 61.10  / 59.42
\\
    NewsES & 58.00 / 55.58  / 51.52& 58.40 / 55.98  / 51.99& 61.23 / 61.53  / 57.01& 61.79 / 61.58  / 57.34& 58.78 / 55.85  / 50.81
\\
    ALPACA & 40.12 / 25.18  / 22.50& 39.62 / 26.77  / 24.14& 41.94 / 27.71  / 22.35& 42.77 / 28.13  / 25.21& 38.92 / 23.54  / 20.10\\
    \bottomrule
    \multicolumn{6}{c}{\textbf{Qwen-2 1.5B}} \\
    GSM8K & 84.23 / 86.60  / 85.15& 84.08 / 86.48  / 84.98& 79.93 / 84.02  / 81.54& 7.82 / 14.80  / 12.37& 86.62 / 87.73  / 85.15
\\
    NewsDE & 64.39 / 63.63  / 62.16& 64.72 / 64.02  / 62.82& 60.77 / 58.30  / 57.15& 5.09 / 11.68  / 11.07& 63.31 / 64.62  / 64.63
\\
    NewsIT & 65.34 / 65.34  / 63.58& 66.54 / 66.74  / 64.90& 64.37 / 63.15  / 61.27& 4.38 / 9.35  / 8.57& 65.98 / 65.43  / 64.27
\\
    NewsES & 65.74 / 66.71  / 65.57& 64.82 / 65.19  / 64.05& 61.97/ 61.39  / 49.41& 5.67 / 14.39  / 12.36& 65.88 / 67.31  / 66.74
\\
    ALPACA & 65.49 / 61.63  / 55.08& 65.78 / 62.00  / 55.50& 57.03 / 55.28  / 49.41& 4.80 / 15.53  / 12.80& 66.82 / 64.31  / 55.19\\
    \bottomrule
    \multicolumn{6}{c}{\textbf{Qwen-2 7B}} \\
    GSM8K & 91.94 / 94.54  / 94.37& 91.80 / 94.27  / 94.17& 57.03 / 55.28  / 49.41& 2.40 / 5.36  / 4.68& 94.33 / 94.98  / 95.00
\\
    NewsDE & 66.39 / 66.15  / 64.72& 66.87 / 66.86 / 65.71& 57.03 / 55.28  / 49.41& 6.59 / 8.68  / 7.88& 68.83 / 67.70  / 64.74
\\
    NewsIT & 68.72 / 69.00  / 67.81& 69.17 / 69.32  / 68.23& 57.03 / 55.28  / 49.41& 6.24 / 9.04  / 8.90& 68.54 / 70.30  / 69.34
\\
    NewsES & 67.96 / 69.49  / 68.80& 67.37 / 68.77  / 68.08& 57.03 / 55.28  / 49.41& 7.26 / 10.21  / 9.73& 68.89 / 69.94  / 70.00
\\
    ALPACA & 66.55 / 63.14  / 57.42& 67.17/ 64.05  / 58.38& 57.03 / 55.28  / 49.41& 6.61 / 12.78  / 11.32& 68.25 / 65.64  / 58.86\\
    \bottomrule
    \multicolumn{6}{c}{\textbf{Llama-3 8B}} \\
    GSM8K & 84.53 / 87.62  / 85.96& 83.21 / 88.20  / 86.73& 59.73 / 56.81  / 52.22& 61.92 / 59.31  / 55.49& 86.67 / 88.72  / 86.54\\
    NewsDE & 63.32 / 61.20  / 56.48& 61.88 / 60.29  / 55.33& 63.50 / 61.69  / 56.32& 65.34 / 63.58 / 58.48& 66.10 / 65.68  / 64.60
\\
    NewsIT & 65.12 / 64.00  / 63.03& 64.92 / 64.37  / 63.45& 66.01 / 64.52  / 63.49& 67.11 / 67.23  / 66.49& 68.63/ 69.08  / 67.39
\\
    NewsES & 59.15 / 55.98  / 52.04& 53.69 / 53.25  / 49.24& 62.48 / 60.84  / 56.36& 60.81 / 58.45  / 54.89& 68.97 / 69.97  / 69.20
\\
    ALPACA & 62.11 / 57.03  / 51.20& 64.15 / 58.91  / 55.22& 37.11 / 23.24  / 19.95& 48.06 / 36.66  / 34.83& 65.84 / 62.83  / 57.11\\
    \bottomrule
    \multicolumn{6}{c}{\textbf{Llama-3.1 8B}} \\
    GSM8K & 84.41 / 87.29  / 85.02& 84.25 / 86.89  / 85.11& 60.36 / 57.60  / 53.31& 62.09 / 58.99  / 55.20& 86.03 / 87.93  / 86.04\\
    NewsDE & 62.85 / 60.54  / 55.85& 61.19 / 59.72  / 54.93& 65.80 / 62.57  / 57.31& 65.93 / 63.98  / 58.92& 66.02 / 66.86  / 65.87
\\
    NewsIT & 65.35 / 64.50  / 63.40& 64.98 / 64.33  / 63.47& 65.12 / 63.73  / 62.75& 67.69 / 67.40  / 66.65& 68.27 / 69.18  / 67.63
\\
    NewsES & 59.31 / 55.94  / 52.08& 54.25 / 53.39  / 49.73& 61.36 / 60.00  / 55.90& 62.10 / 59.63  / 55.87& 69.05 / 70.29  / 69.25
\\
    ALPACA & 61.03 / 58.92  / 51.09& 61.93 / 60.21  / 52.33& 37.89 / 23.94  / 20.89& 48.67 / 35.96  / 33.87& 67.10 / 64.11  / 58.77\\
    \bottomrule
    \multicolumn{6}{c}{\textbf{Avg.}} \\
    GSM8K & 80.69 / 82.41 / 80.24& 80.84 / 82.80 / 80.94& 67.84 / 68.12  / 64.42& 38.89 / 39.44  / 36.32& 82.49 / 83.12 / 80.73\\
    NewsDE & 64.08 / 62.82  / 59.28& 63.64 / 62.69 / 59.11& 63.07 / 61.41 / 57.56& 41.16 / 42.05  / 38.61& 65.25 / 65.11  / 62.25
\\
    NewsIT & 66.01 / 65.42  / 64.16& 66.33 / 66.09  / 64.93& 65.03/ 63.77 / 62.47& 42.10 / 43.34 / 42.57& 66.67 / 67.02  / 65.61
\\
    NewsES & 62.03 / 60.74  / 58.00& 59.70 /  59.31 / 56.62& 62.15 / 61.49  / 58.48& 39.53 / 40.85  / 38.04& 66.31 / 66.67  / 65.20
\\
    ALPACA & 59.05 / 53.17 / 47.45& 59.73 / 54.38 / 49.11& 49.13 / 39.55  / 34.61& 30.18 / 25.81  / 23.61& 61.39 / 56.09  / 50.00\\
    \bottomrule
    \end{tabular}}
   \caption{EXP1-QA BLEU/Rouge1/RougeL}
    \label{tab:EXP1-QA}
\end{table*}

\begin{table*}[tb]
    \centering
    \small
    \resizebox{\textwidth}{!}{
    \begin{tabular}{lccccc}
    \toprule
    \textbf{Task} & \textbf{LoRA} & \textbf{DoRA} & \textbf{CUR-LoRA} & \textbf{ILoRA} & \textbf{SECURA} \\
    \midrule
    \multicolumn{6}{c}{\textbf{Gemma-2 2B}} \\
    AbsNarr
& 00.09 / 00.24  / 00.24& 00.17 / 00.11  / 00.11& 19.91 / 20.13  / 16.41& 46.01 / 49.41 / 42.03& 64.18 / 63.77 / 50.89
\\
    DisflQA
& 00.10 / 00.58 / 00.57& 00.19 / 00.52  / 00.50& 08.78 / 11.92  / 09.44& 65.92 / 66.79 / 60.03& 65.16 / 65.00 / 52.01
\\
    LangID
& 00.30 / 00.52  / 00.50& 00.06 / 00.20  / 00.20& 31.57 / 39.89  / 32.39& 66.61 / 67.49 / 60.81& 65.16 / 64.88 / 52.19
\\
    LogDeduc
& 03.14 / 04.25  / 03.71& 02.01 / 00.26  / 00.25& 17.23 / 22.95  / 17.70& 68.95 / 70.12 / 63.87& 66.86 / 67.70 / 55.27
\\
    ObjCount
& 30.59 / 30.54  / 21.41& 39.57 / 42.87  / 32.58& 49.18 / 52.11  / 40.85& 68.59 / 69.04 / 62.03& 62.37 / 63.15 / 50.86
\\
    PlayDiag
& 00.07 / 01.23  / 01.21& 00.15 / 01.73  / 01.63& 16.33 / 19.69  / 14.99& 60.30 / 62.69 / 56.41& 63.56 / 64.47 / 51.53
\\
    ColorReason
& 02.96 / 06.16  / 05.10& 00.59 / 01.36  / 01.26& 23.94 / 30.21  / 22.94& 62.85 / 63.64 / 57.69& 64.93 / 64.64 / 51.72
\\
    TrackObj& 00.12 / 01.19  / 01.17& 00.16 / 00.38  / 00.36& 01.10 / 03.73  / 03.34& 41.25 / 42.73 / 35.41& 64.28 / 64.67 / 51.79\\
    \bottomrule
    \multicolumn{6}{c}{\textbf{Qwen-2 1.5B}} \\
    AbsNarr
& 03.75 / 03.72 / 03.31& 06.35 / 05.69 / 04.82& 00.28 / 00.11 / 00.11& 00.82 / 00.39 / 00.36& 75.07 / 75.83 / 70.80
\\
    DisflQA
& 29.04 / 37.23 / 29.42& 47.51 / 48.82 / 40.14& 34.54 / 38.93 / 30.47& 00.32 / 01.03 / 01.00& 75.83 / 76.99 / 72.15
\\
    LangID
& 00.50 / 00.00 / 00.00& 00.26 / 00.00 / 00.00& 00.24 / 00.19 / 00.19& 00.60 / 01.70 / 01.63& 76.75 / 76.98 / 72.44
\\
    LogDeduc
& 00.26 / 00.34 / 00.34& 00.29 / 00.19 / 00.19& 00.27 / 00.87 / 00.80& 01.29 / 01.75 / 01.52& 76.92 / 77.21 / 71.87
\\
    ObjCount
& 01.12 / 00.14 / 00.14& 00.84 / 00.00 / 00.00& 06.99 / 08.61 / 07.43& 02.43 / 04.65 / 03.77& 73.31 / 75.28 / 70.46
\\
    PlayDiag
& 00.18 / 00.00 / 00.00& 00.31 / 00.00 / 00.00& 00.27 / 00.01 / 00.01& 00.29 / 00.72 / 00.70& 74.98 / 76.57 / 72.04
\\
    ColorReason
& 00.35 / 00.00 / 00.00& 00.25 / 00.04 / 00.04& 04.81 / 08.01 / 06.27& 00.29 / 00.02 / 00.02& 75.01 / 75.78 / 70.83
\\
    TrackObj& 00.62 / 00.00 / 00.00& 05.11 / 08.39 / 07.84& 00.42 / 00.72 / 00.66& 00.48 / 00.62 / 00.60& 74.71 / 76.28 / 71.37\\
    \bottomrule
    \multicolumn{6}{c}{\textbf{Qwen-2 7B}} \\
    AbsNarr
& 00.28 / 00.05 / 00.05
& 00.22 / 00.12 / 00.12
& 00.67 / 00.35 / 00.35
& 00.18 / 00.00 / 00.00& 78.18 / 79.74 / 75.87
\\
    DisflQA
& 00.29 / 00.16 / 00.16
& 41.88 / 44.69 / 34.05
& 01.00 / 00.59 / 00.54
& 00.28 / 00.00 / 00.00& 78.46 / 80.25 / 76.70
\\
    LangID
& 00.68 / 00.00 / 00.00
& 00.65 / 00.00 / 00.00
& 00.29 / 00.01 / 00.01
& 00.68 / 00.00 / 00.00& 78.58 / 80.88 / 77.13
\\
    LogDeduc
& 00.50 / 00.85 / 00.85
& 00.30 / 00.14 / 00.14
& 49.70 / 54.73 / 49.87
& 00.52 / 00.22 / 00.22& 75.41 / 77.01 / 73.15
\\
    ObjCount
& 00.95 / 00.00 / 00.00
& 01.32 / 00.10 / 00.10
& 00.68 / 00.05 / 00.05
& 00.88 / 00.00 / 00.00& 77.71 / 80.64 / 76.56
\\
    PlayDiag
& 00.21 / 00.00 / 00.00
& 00.63 / 00.00 / 00.00
& 00.55 / 00.56 / 00.56
& 00.30 / 00.03 / 00.03& 77.39 / 80.31 / 76.62
\\
    ColorReason
& 00.65 / 00.02 / 00.02
& 00.20 / 00.00 / 00.00
& 00.30 / 00.00 / 00.00
& 00.22 / 00.02 / 00.02& 78.09 / 79.62 / 75.72
\\
    TrackObj& 00.38 / 00.02 / 00.02& 00.01 / 00.00 / 00.00& 08.70 / 07.30 / 06.39& 00.27 / 00.09 / 00.09& 78.06 / 80.36 / 76.74\\
    \bottomrule
    \multicolumn{6}{c}{\textbf{Llama-3 8B}} \\
    AbsNarr
& 00.15 / 00.06 / 00.06& 00.07 / 00.01 / 00.01& 01.90 / 03.47 / 03.13& 00.08 / 00.30 / 00.30& 75.32 / 76.84 / 70.77
\\
    DisflQA
& 00.08 / 00.15 / 00.15& 00.05 / 00.44 / 00.44& 00.89 / 04.11 / 03.93& 00.11 / 01.78 / 01.71& 74.32 / 76.77 / 69.94
\\
    LangID
& 00.14 / 00.03 / 00.03& 16.73 / 21.59 / 18.88& 01.38 / 07.69 / 07.23& 00.08 / 00.01 / 00.01& 74.98 / 76.66 / 70.29
\\
    LogDeduc
& 00.88 / 00.07 / 00.07& 00.52 / 00.06 / 00.06& 10.75 / 12.65 / 09.71& 59.44 / 62.69 / 55.91& 75.30 / 76.90 / 71.00
\\
    ObjCount
& 09.13 / 06.23 / 05.57& 00.51 / 00.03 / 00.03& 34.71 / 38.38 / 26.96& 74.02 / 75.15 / 68.88& 73.36 / 74.92 / 68.97
\\
    PlayDiag
& 00.07 / 00.05 / 00.05& 00.20 / 02.25 / 02.25& 03.50 / 08.47 / 07.40& 62.97 / 65.34 / 57.40& 75.31 / 77.47 / 71.76
\\
    ColorReason
& 00.19 / 01.09 / 00.97& 00.11 / 00.17 / 00.17& 06.65 / 10.82 / 08.78& 66.22 / 68.48 / 59.60& 75.79 / 76.93 / 70.36
\\
    TrackObj& 00.20 / 00.02 / 00.02& 00.13 / 00.08 / 00.07& 00.08 / 00.18 / 00.18& 69.13 / 71.03 / 63.49& 74.80 / 76.94 / 70.88\\
    \bottomrule
    \multicolumn{6}{c}{\textbf{Llama-3.1 8B}} \\
    AbsNarr
& 00.01 / 00.00 / 00.00& 00.10 / 00.00 / 00.00& 00.22 / 00.04 / 00.04
& 00.07 / 00.03 / 00.03
& 75.61 / 77.14 / 71.44
\\
    DisflQA
& 00.05 / 00.46 / 00.46& 32.60 / 38.93 / 28.23& 01.69 / 05.85 / 05.58
& 68.98 / 70.84 / 63.32
& 74.45 / 77.02 / 70.10
\\
    LangID
& 00.09 / 02.16 / 02.14& 00.10 / 00.34 / 00.34& 00.25 / 01.55 / 01.55
& 56.69 / 58.30 / 50.32
& 74.43 / 75.85 / 69.13
\\
    LogDeduc
& 00.18 / 00.05 / 00.05& 00.09 / 00.00 / 00.00& 11.53 / 14.63 / 11.81
& 00.67 / 00.08 / 00.08
& 75.84 / 77.74 / 71.77
\\
    ObjCount
& 06.98 / 04.58 / 04.20& 00.29 / 00.00 / 00.00& 18.15 / 21.50 / 16.06
& 70.13 / 72.79 / 66.51
& 74.22 / 76.07 / 69.96
\\
    PlayDiag
& 00.07 / 00.09 / 00.09& 00.21 / 00.67 / 00.67& 02.33 / 07.99 / 06.87
& 47.87 / 51.97 / 42.48
& 75.13 / 77.56 / 71.74
\\
    ColorReason
& 00.05 / 00.03 / 00.03& 00.10 / 00.00 / 00.00& 00.75 / 02.58 / 02.30
& 71.29 / 72.11 / 65.31
& 76.65 / 77.63 / 71.04
\\
    TrackObj& 00.04 / 00.09 / 00.09& 00.10 / 00.15 / 00.15& 01.06 / 07.59 / 06.86& 00.16 / 00.62 / 00.59& 74.90 / 77.09 / 71.06\\
    \bottomrule
    \multicolumn{6}{c}{\textbf{Avg.}} \\
    AbsNarr
& 00.85 / 00.81 / 00.73
& 01.38 / 01.19 / 01.01& 04.60 / 04.82 / 04.01& 09.43 / 10.03 / 08.54& 73.67 / 74.66 / 67.95
\\
    DisflQA
& 05.91 / 07.72 / 06.15
& 24.44 / 26.68 / 20.67& 09.38 / 12.28 / 09.99& 27.12 / 28.09 / 25.21& 73.65 / 75.20 / 68.18
\\
    LangID
& 00.34 / 00.54 / 00.53
& 03.56 / 04.42 / 03.88& 06.74 / 09.86 / 08.27& 24.93 / 25.50 / 22.55& 73.98 / 75.05 / 68.24
\\
    LogDeduc
& 00.99 / 01.11 / 01.00
& 00.64 / 00.13 / 00.13& 17.89 / 21.17 / 17.98& 26.18 / 26.97 / 24.32& 74.07 / 75.31 / 68.61
\\
    ObjCount
& 09.75 / 08.30 / 06.26
& 08.51 / 08.60 / 06.54& 21.94 / 24.13 / 18.27& 43.21 / 44.32 / 40.24& 72.19 / 74.01 / 67.36
\\
    PlayDiag
& 00.12 / 00.27 / 00.27
& 00.30 / 00.93 / 00.91& 04.59 / 07.35 / 05.96& 34.35 / 36.15 / 31.40& 73.27 / 75.28 / 68.74
\\
    ColorReason
& 00.84 / 01.46 / 01.22
& 00.25 / 00.31 / 00.29& 07.29 / 10.32 / 08.06& 40.17 / 40.85 / 36.53& 74.09 / 74.92 / 67.93
\\
    TrackObj& 00.27 / 00.26 / 00.26& 01.10 / 01.80 / 01.68& 02.27 / 03.90 / 03.48& 22.26 / 23.02 / 20.04& 73.35 / 75.07 / 68.37\\
    \bottomrule
    \end{tabular}}
   \caption{EXP2-Forgetting BLEU/Rouge1/RougeL}
    \label{tab:EXP2-Forgetting}
\end{table*}

\begin{table*}[h]
\centering
\small
\begin{tabular}{lccccc}
\toprule
\textbf{Datasets} & \textbf{SEQ} & \textbf{EWC}& \textbf{ER} & \textbf{I-LORA} & \textbf{SECURA} \\
\midrule
AbsNarr-1k
& 73.28 / 75.55 / 68.58
& 75.39 / 58.84 / 59.90
& 67.77 / 75.14 / 69.68
& 75.57 / 76.75 / 70.81
& 78.40 / 79.69 / 75.94
\\

AbsNarr-2k
& 
66.26 / 64.33 / 61.01
& 70.19 / 58.05 / 55.68
& 76.42 / 77.77 / 70.94
& 
75.25 / 76.69 / 71.20
&

78.01 / 79.28 / 75.86
\\
DisflQA-1k
& 
64.90 / 65.93 / 60.63
& 67.96 / 67.08 / 60.41
& 72.77 / 75.70 / 73.73
& 74.96 / 75.95 / 70.25
& 
77.96 / 78.94 / 74.97
\\

DisflQA-2k
& 
54.93 / 47.38 / 42.97
& 62.43 / 68.62 / 67.01
& 71.68 / 73.71 / 70.32
& 
74.93 / 76.06 / 70.26
&

76.46 / 77.96 / 73.85
\\
CNNDM-1k
& 46.59 / 48.36 / 37.78
& 68.43 / 62.65 / 62.91
& 69.43 / 77.45 / 70.41
& 74.92 / 75.74 / 69.97
& 78.55 / 79.21 / 75.39
\\

CNNDM-2k
& 
37.53 / 31.12 / 24.50
& 67.45 / 58.24 / 64.16
& 73.04 / 69.97 / 73.83
& 
74.35 / 75.85 / 70.13
&

78.29 / 78.87 / 74.77
\\

LangID-1k
& 
28.68 / 24.94 / 25.06
& 68.48 / 59.60 / 66.68
& 78.57 / 78.31 / 72.06
& 74.29 / 75.70 / 69.92
& 
76.43 / 77.76 / 73.87
\\

LangID-2k
& 
25.54 / 18.47 / 19.13
& 74.32 / 64.99 / 55.31
& 74.75 / 74.98 / 70.12
& 
74.97 / 76.12 / 70.23
&

77.05 / 78.35 / 74.55
\\
LingPuzz-1k
& 21.64 / 15.06 / 13.97
& 62.64 / 55.22 / 59.36
& 72.33 / 69.01 / 68.93
& 70.89 / 71.89 / 66.51
& 77.44 / 78.50 / 74.29
\\

LingPuzz-2k
& 
17.24 / 14.78 / 14.79
& 70.94 / 56.87 / 59.37
& 69.88 / 77.20 / 67.92
& 
71.32 / 72.19 / 66.55
&

77.84 / 79.69 / 75.65
\\

LogDeduc-1k
& 
12.28 / 21.61 / 10.62
& 74.88 / 68.39 / 55.98
& 68.73 / 75.06 / 69.96
& 67.38 / 69.78 / 64.14
& 
76.11 / 77.66 / 73.49
\\

LogDeduc-2k
& 
12.18 / 15.29 / 11.70
& 64.50 / 56.40 / 65.20
& 74.25 / 72.63 / 65.04
& 
68.26 / 69.99 / 64.19
&

75.81 / 77.55 / 73.58
\\
NewsDE-1k
& 18.56 / 19.40 / 10.89
& 66.71 / 72.97 / 53.78
& 75.47 / 76.34 / 71.89
& 69.49 / 70.59 / 64.94
& 77.17 / 78.14 / 73.93
\\

NewsDE-2k
& 
12.48 / 16.91 / 15.00
& 73.97 / 55.39 / 61.35
& 73.69 / 72.29 / 66.30
& 
70.12 / 70.68 / 64.76
&

77.29 / 78.04 / 73.90
\\

NewsES-1k
& 
13.14 / 13.92 / 09.41
& 76.47 / 65.25 / 66.89
& 70.59 / 76.85 / 70.16
& 71.13 / 72.11 / 66.38
& 
78.52 / 79.12 / 75.09
\\

NewsES-2k
& 
17.57 / 18.70 / 15.63
& 74.44 / 71.62 / 56.97
& 75.49 / 77.96 / 71.54
& 
70.37 / 71.27 / 65.09
&

77.32 / 78.13 / 74.06
\\
NewsIT-1k
& 16.79 / 15.78 / 15.28
& 70.15 / 60.99 / 65.53
& 73.05 / 69.87 / 69.61
& 69.73 / 71.28 / 64.91
& 77.04 / 77.81 / 73.60
\\

NewsIT-2k
& 
16.21 / 13.61 / 14.03
& 59.92 / 64.77 / 56.84
& 68.92 / 74.91 / 70.74
& 
69.79 / 70.75 / 64.27
&

77.19 / 78.18 / 74.08
\\

ObjCount-1k
& 
20.04 / 19.88 / 12.88
& 58.89 / 54.79 / 59.17
& 73.25 / 75.17 / 73.22
& 70.01 / 71.12 / 64.56
& 
75.87 / 76.80 / 72.09
\\

ObjCount-2k
& 
16.21 / 14.24 / 10.50
& 61.86 / 70.82 / 53.73
& 76.96 / 74.24 / 70.42
& 
69.75 / 71.01 / 64.33
&

75.23 / 76.19 / 71.88
\\
PlayDiag-1k
& 20.20 / 20.68 / 14.39
& 66.16 / 57.02 / 61.08
& 76.24 / 74.59 / 73.92
& 69.94 / 71.24 / 64.15
& 73.15 / 73.39 / 68.69
\\

PlayDiag-2k
& 
16.54 / 14.07 / 09.09
& 68.42 / 67.72 / 66.00
& 71.71 / 76.02 / 69.82
& 
70.16 / 71.32 / 64.23
&

73.10 / 74.33 / 69.90
\\

QuesSel-1k
& 
16.48 / 18.08 / 08.91
& 76.35 / 64.25 / 56.73
& 72.41 / 78.06 / 71.14
& 66.44 / 67.81 / 61.11
& 
74.90 / 76.64 / 72.43
\\

QuesSel-2k
& 
11.53 / 13.89 / 13.83
& 71.47 / 59.96 / 57.23
& 77.98 / 71.05 / 68.51
& 
66.31 / 67.98 / 61.34
&

75.19 / 76.92 / 72.87
\\
ColorReason-1k
& 20.86 / 20.52 / 08.99
& 62.26 / 60.83 / 56.48
& 68.15 / 70.82 / 75.76
& 61.76 / 64.05 / 57.77
& 76.84 / 77.60 / 73.47
\\

ColorReason-2k
& 
19.30 / 21.79 / 12.51
& 68.17 / 63.93 / 55.37
& 75.77 / 71.09 / 68.59
& 
61.77 / 63.61 / 57.11
&

76.59 / 77.66 / 73.61
\\

StratQA-1k
& 
14.51 / 20.64 / 09.89
& 73.60 / 55.71 / 59.90
& 73.58 / 74.30 / 65.52
& 69.63 / 70.62 / 63.77
& 
78.36 / 79.07 / 74.86
\\

StratQA-2k
& 
16.54 / 16.59 / 16.79
& 64.61 / 57.61 / 66.30
& 77.37 / 71.37 / 72.18
& 
71.34 / 71.79 / 64.86
&

79.14 / 79.83 / 75.58
\\
TrackObj-1k
& 18.87 / 18.21 / 13.45
& 74.40 / 64.98 / 61.91
& 67.89 / 74.82 / 70.75
& 69.08 / 70.41 / 63.87
& 76.03 / 77.06 / 72.91
\\

TrackObj-2k
& 
19.24 / 17.74 / 16.63
& 68.89 / 65.52 / 61.10
& 72.40 / 73.86 / 69.43
& 
69.60 / 70.37 / 63.30
&

75.55 / 76.38 / 72.32
\\

ALPACA-1k
& 
19.74 / 14.36 / 15.40
& 67.77 / 63.03 / 62.00
& 73.81 / 74.09 / 67.03
& 69.07 / 70.09 / 62.71
& 
78.32 / 78.97 / 74.92
\\

ALPACA-2k& 
18.90 / 17.15 / 15.64& 64.89 / 62.35 / 61.67
& 70.38 / 69.34 / 69.43
& 
69.87 / 70.60 / 63.67&

78.53 / 79.79 / 75.93\\
\bottomrule
\end{tabular}
\caption{EXP3 Qwen-2 7B}
\label{tab:EXP3 Qwen-2 7B}
\end{table*}

\begin{table*}[h]
\centering
\small
\begin{tabular}{lccccc}
\toprule
\textbf{Datasets} & \textbf{SEQ}& \textbf{EWC}& \textbf{ER} & \textbf{I-LORA} & \textbf{SECURA} \\
\midrule
AbsNarr-1k& 69.18 / 63.32 / 54.88
& 65.54 / 61.29 / 58.09
& 70.07 / 73.84 / 74.78
& 75.57 / 76.75 / 70.81
& 76.95 / 78.11 / 71.80
\\

AbsNarr-2k& 
59.20 / 47.47 / 55.10
& 62.06 / 58.39 / 62.74
& 67.46 / 71.10 / 72.56
& 
75.25 / 76.69 / 71.20
&

75.70 / 76.96 / 70.62
\\
DisflQA-1k& 
55.71 / 46.66 / 47.68
& 63.52 / 58.81 / 58.65
& 68.45 / 75.35 / 68.83
& 74.96 / 75.95 / 70.25
& 
75.58 / 76.86 / 70.58
\\

DisflQA-2k& 
54.12 / 41.02 / 51.27
& 60.74 / 59.32 / 63.52
& 73.34 / 74.68 / 67.86
& 
74.93 / 76.06 / 70.26
&

74.83 / 76.61 / 70.34
\\
CNNDM-1k& 49.91 / 40.61 / 45.57
& 68.43 / 63.42 / 63.54
& 73.78 / 75.84 / 67.71
& 74.92 / 75.74 / 69.97
& 76.36 / 78.03 / 71.65
\\

CNNDM-2k& 
45.65 / 34.81 / 43.70
& 67.03 / 60.14 / 57.00
& 73.42 / 74.05 / 70.98
& 
74.35 / 75.85 / 70.13
&

76.06 / 77.52 / 71.24
\\

LangID-1k& 
41.22 / 33.40 / 28.07
& 64.08 / 63.61 / 58.88
& 73.27 / 74.73 / 69.55
& 74.29 / 75.70 / 69.92
& 
73.83 / 76.12 / 69.80
\\

LangID-2k& 
37.41 / 33.45 / 26.26
& 68.13 / 58.36 / 57.56
& 74.64 / 79.49 / 66.13
& 
74.97 / 76.12 / 70.23
&

73.55 / 75.70 / 69.52
\\
LingPuzz-1k& 37.15 / 29.39 / 30.99
& 60.18 / 67.05 / 56.20
& 70.67 / 78.64 / 69.14
& 70.89 / 71.89 / 66.51
& 75.77 / 76.93 / 70.69
\\

LingPuzz-2k& 
28.57 / 24.65 / 24.37
& 59.40 / 58.79 / 56.76
& 72.65 / 73.37 / 71.63
& 
71.32 / 72.19 / 66.55
&

75.43 / 76.46 / 70.42
\\

LogDeduc-1k& 
26.00 / 13.26 / 16.88
& 67.22 / 58.51 / 58.65
& 76.15 / 78.50 / 68.74
& 67.38 / 69.78 / 64.14
& 
75.48 / 76.63 / 70.70
\\

LogDeduc-2k& 
19.16 / 07.08 / 16.08
& 64.65 / 60.56 / 56.30
& 71.59 / 71.41 / 70.79
& 
68.26 / 69.99 / 64.19
&

76.25 / 77.13 / 70.94
\\
NewsDE-1k& 14.37 / 10.09 / 15.45
& 64.15 / 60.38 / 60.14
& 73.55 / 75.26 / 70.54
& 69.49 / 70.59 / 64.94
& 75.92 / 76.90 / 70.82
\\

NewsDE-2k& 
15.66 / 06.82 / 13.88
& 59.85 / 61.09 / 57.85
& 68.35 / 79.48 / 66.44
& 
70.12 / 70.68 / 64.76
&

76.35 / 77.32 / 71.30
\\

NewsES-1k& 
14.11 / 12.49 / 06.47
& 59.54 / 64.15 / 57.66
& 70.96 / 73.53 / 69.15
& 71.13 / 72.11 / 66.38
& 
76.92 / 78.02 / 71.96
\\

NewsES-2k& 
17.65 / 09.82 / 16.77
& 67.33 / 64.66 / 55.14
& 68.49 / 70.50 / 65.03
& 
70.37 / 71.27 / 65.09
&

75.87 / 77.30 / 71.52
\\
NewsIT-1k& 12.58 / 09.18 / 17.16
& 68.69 / 64.00 / 62.40
& 78.12 / 75.56 / 68.42
& 69.73 / 71.28 / 64.91
& 76.25 / 77.36 / 71.19
\\

NewsIT-2k& 
10.55 / 05.95 / 08.20
& 69.35 / 60.45 / 62.54
& 68.89 / 74.56 / 69.71
& 
69.79 / 70.75 / 64.27
&

76.37 / 77.48 / 71.27
\\

ObjCount-1k& 
16.21 / 06.25 / 09.35
& 58.48 / 67.04 / 54.52
& 72.91 / 72.87 / 68.49
& 70.01 / 71.12 / 64.56
& 
75.84 / 77.38 / 71.64
\\

ObjCount-2k& 
15.88 / 07.88 / 08.46
& 62.76 / 60.67 / 57.22
& 72.70 / 74.70 / 73.83
& 
69.75 / 71.01 / 64.33
&

74.11 / 76.30 / 70.45
\\
PlayDiag-1k& 18.61 / 11.20 / 07.12
& 65.28 / 64.65 / 55.17
& 72.49 / 79.93 / 67.73
& 69.94 / 71.24 / 64.15
& 75.47 / 76.68 / 70.63
\\

PlayDiag-2k& 
12.40 / 06.14 / 15.20
& 67.52 / 60.96 / 57.17
& 72.43 / 75.20 / 65.62
& 
70.16 / 71.32 / 64.23
&

74.67 / 76.26 / 70.04
\\

QuesSel-1k& 
12.86 / 12.78 / 08.42
& 68.95 / 57.29 / 56.37
& 68.78 / 70.81 / 65.90
& 66.44 / 67.81 / 61.11
& 
75.10 / 76.61 / 70.44
\\

QuesSel-2k& 
15.07 / 06.14 / 12.65
& 69.17 / 60.35 / 64.00
& 68.59 / 78.57 / 75.98
& 
66.31 / 67.98 / 61.34
&

75.08 / 76.64 / 70.40
\\
ColorReason-1k& 13.96 / 09.85 / 09.35
& 59.17 / 64.56 / 57.21
& 72.10 / 69.43 / 73.29
& 61.76 / 64.05 / 57.77
& 74.60 / 76.52 / 70.37
\\

ColorReason-2k& 
13.76 / 11.97 / 11.27
& 60.30 / 58.24 / 58.42
& 73.58 / 71.75 / 76.00
& 
61.77 / 63.61 / 57.11
&

75.05 / 76.47 / 70.53
\\

StratQA-1k& 
16.80 / 06.01 / 09.41
& 67.17 / 65.51 / 66.36
& 72.31 / 78.95 / 67.90
& 69.63 / 70.62 / 63.77
& 
75.76 / 77.33 / 71.64
\\

StratQA-2k& 
15.74 / 10.51 / 15.28
& 61.96 / 63.57 / 53.82
& 71.80 / 73.93 / 66.00
& 
71.34 / 71.79 / 64.86
&

76.32 / 77.51 / 71.45
\\
TrackObj-1k& 17.79 / 14.76 / 08.23
& 70.91 / 57.14 / 56.88
& 73.40 / 77.29 / 73.51
& 69.08 / 70.41 / 63.87
& 76.99 / 77.77 / 71.78
\\

TrackObj-2k& 
12.14 / 08.31 / 17.06
& 58.71 / 55.59 / 64.89
& 69.81 / 70.79 / 70.75
& 
69.60 / 70.37 / 63.30
&

77.38 / 78.09 / 72.19
\\

ALPACA-1k& 
12.90 / 09.89 / 06.41
& 66.69 / 60.25 / 63.42
& 74.06 / 76.02 / 69.19
& 69.07 / 70.09 / 62.71
& 
76.96 / 77.55 / 71.54
\\

ALPACA-2k& 
13.25 / 15.62 / 12.66
& 66.88 / 59.11 / 66.27
& 68.09 / 80.43 / 73.90
& 
69.87 / 70.60 / 63.67
&

75.12 / 76.38 / 70.43
\\
\bottomrule
\end{tabular}
\caption{EXP3 Llama3 8B}
\label{tab:EXP3 Llama3 8B}
\end{table*}

%% file: sections/secura_algorithm.tex
\SetAlFnt{\small}

\begin{algorithm}[htbp]  
\caption{SECURA Methodology}
\label{alg:secura}
\SetAlgoLined
\DontPrintSemicolon
\textbf{Input}: Base weights $\mathcal{W}_{\text{base}}$, hyperparameters $r, m, \epsilon, \text{Scale}$ \\
\textbf{Output}: Updated weights $\mathcal{W}_{\text{updated}}$

\underline{\textbf{Step 1: CABR-LoRA Decomposition}} \\
1. \textit{Initialize CABR matrices}: 
\quad $\mathcal{W}_{\text{frozen}} = \mathcal{U}_{\text{SVD}} \mathcal{S}_{\text{SVD}} \mathcal{V}_{\text{SVD}}^{\top}.$ \tcp*{SVD of base weights} 
\quad $\mathcal{W}_A \gets \mathcal{U}_{\text{SVD}}^{(r,:)} \mathcal{S}_{\text{SVD}} \mathcal{V}_{\text{SVD}}^{(:,m) \top}$ \tcp*{Truncated SVD for $A$} 
\quad $\mathcal{W}_B \gets \text{Zeros}(m \times r)$ \tcp*{Zero-initialized $B$} 
\quad $\mathcal{W}_{\text{CABR}} \gets C \cdot \mathcal{W}_A \cdot \mathcal{W}_B \cdot R$ \tcp*{CABR reconstruction}

\underline{\textbf{Step 2: S-MagNorm Normalization}} \\
2. \textit{Merge weights}: \\
\quad $\mathcal{W}_{\text{SECURAMerged}} \gets \mathcal{W}_{\text{CABR}} + \mathcal{W}_{\text{base}}$ \\
3. \textit{Compute magnitude ratio}: \\
\quad $\mathcal{M}_{\text{Mag}} \gets \left|\frac{\mathcal{W}_{\text{SECURAMerged}}}{\mathcal{W}_{\text{base}} + \epsilon}\right|$ \tcp*{Avoid division by zero} 
4. \textit{Normalize and scale}: \\
\quad $\mathcal{M}_{\text{norm}} \gets \left[\frac{\mathcal{M}_{\text{Mag}}}{\max(\mathcal{M}_{\text{Mag}}) + \epsilon} - 0.5\right] \cdot \text{Scale}$ \\
5. \textit{Apply Sigmoid constraint}: \\
\quad $\mathcal{M}_{\text{Res}} \gets 2 - \sigma(\mathcal{M}_{\text{norm}})$ \tcp*{$\sigma$: Sigmoid function} 
6. \textit{Adjust updates}: \\
\quad $\mathcal{W}_{\text{updated}} \gets \mathcal{W}_{\text{SECURAMerged}} / \mathcal{M}_{\text{Res}}$ \tcp*{Element-wise division}

\underline{\textbf{Step 3: Merge Strategy (M1/M2)}} \\
7. \If{M1 (Deep Merge)}{
    \quad $\mathcal{W}_{\text{updated}} \gets C A_{\text{former}} B_{\text{train}} R + \mathcal{W}_{\text{base}}$ \tcp*{Update base weights}
}
\ElseIf{M2 (Shallow Merge)}{
    \quad $\mathcal{W}_{\text{mergedBase}} \gets C A_{\text{frozen}} B_{\text{accum}} R + \mathcal{W}_{\text{base}}$ \\
    \quad $\mathcal{W}_{\text{merged}} \gets \mathcal{W}_{\text{mergedBase}} + C A_{\text{train}} B_{\text{reset}} R$ \tcp*{Preserve $\mathcal{W}_{\text{basic}}$}
}
\Return $\mathcal{W}_{\text{updated}}$
\end{algorithm}